\documentclass[cameraready]{Interspeech}
\title{Context Biasing for Pronunciation-Orthography Mismatch\\in Automatic Speech Recognition}

\author[affiliation={1}]{Christian}{Huber}
\author[affiliation={2}]{Alexander}{Waibel}
\address{
    $^1$ Interactive Systems Lab, Karlsruhe Institute of Technology, Karlsruhe, Germany\\
    $^2$ Interactive Systems Lab, Carnegie Mellon University, Pittsburgh PA, USA
}

\email{christian.huber@kit.edu, alexander.waibel@cmu.edu}


\usepackage{comment}

\usepackage{cite}
\usepackage{amsmath,amssymb,amsfonts}
\usepackage{algorithmic}
\usepackage{graphicx}
\usepackage{textcomp}
\usepackage{xcolor}

\def\BibTeX{{\rm B\kern-.05em{\sc i\kern-.025em b}\kern-.08em
    T\kern-.1667em\lower.7ex\hbox{E}\kern-.125emX}}

\usepackage{makecell}
\usepackage{multirow}

\usepackage{hyperref} %[colorlinks=true, linkcolor=blue, urlcolor=red, citecolor=green]

\DeclareMathOperator*{\argmax}{argmax}

\usepackage[normalem]{ulem}

\usepackage{pifont}

\usepackage{diagbox}

\begin{document}

\maketitle

\begin{abstract}
Neural sequence-to-sequence systems deliver state-of-the-art performance for automatic speech recognition.
When using appropriate modeling units, e.g., byte-pair encoding, these systems are in principle open vocabulary systems.
In practice, however, they often fail to recognize words not seen during training, e.g., named entities, acronyms, or domain-specific special words.
To address this problem, many context biasing methods have been proposed;
however, these methods may still struggle when they are unable to relate audio and corresponding text, e.g., in case of a pronunciation-orthography mismatch.
We propose a method where corrections of substitution errors can be used to improve the recognition accuracy of such challenging words.
Users can add corrections on the fly during inference.
We show that with this method we get a relative improvement in biased word error rate between 22\% and 34\% compared to a text-based replacement method, while maintaining the overall performance.
\end{abstract}

\section{Introduction}

Up until a few years ago automatic speech recognition (ASR) systems were implemented as Bayes classifiers in order to search for the word sequence $\hat{Y}$, among all possible word sequences $Y$, with the highest posterior probability given a sequence of feature vectors $X$ which is the result of pre-processing the acoustic signal to be recognized:
\begin{align}
    \hat{Y}&=\argmax_Y P(Y|X)\notag\\
    &=\argmax_Y P(X|Y)P(Y)
    \label{eq:bayes}
\end{align}
In the context of ASR $P(X|Y)$ is called the acoustic model, $P(Y)$ the language model.
The space of allowed word sequences to search among was usually defined by a list of words, the vocabulary, of which permissible word sequences could be composed.
Words that were not in the vocabulary could not be recognized.
In turn this means that by adding words to the vocabulary and appropriate probabilities to the language model, previously unknown words could be 
added to the ASR system.
%easily added to the ASR system manually or even automatically.

\begin{figure}[t!]
  \centering
  %\boxed{
  %\includegraphics[page=5,trim=2.4cm 0.5cm 16.0cm 0.3cm,clip,width=1.0\linewidth]{images/figure.pdf}
  \includegraphics[page=1,trim=5.6cm 15.9cm 5.6cm 1.8cm,clip,width=1.0\linewidth]{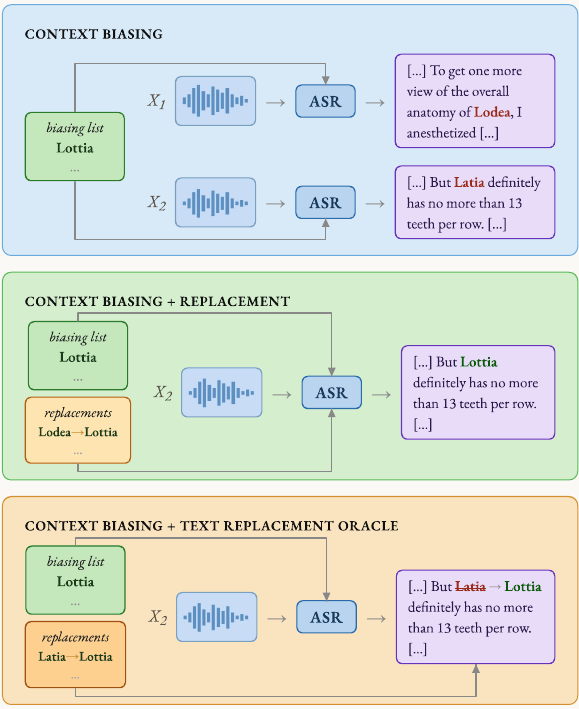}
  %}
  \caption{Approaches:
  Top: Inference of the baseline context biasing ASR model for two utterances containing the same named entity "Lottia" in the reference transcript; the model failed to recognize the named entity.
  Middle: Approach context biasing + replacement; the context biasing list contains the wrongly recognized word "Lodea" from $another$ utterance mapped to "Lottia" (for details how the model uses this see Section \ref{sec:method}). 
  Bottom: Approach context biasing + text replacement oracle for comparison; the replacement "Latia" mapped to "Lottia" from the $same$ utterance is used.}
  \label{fig:approaches}
\end{figure}

In contrast, for neural end-to-end trained ASR systems \cite{vaswani2017attention, pham2019very,radford2023robust}, this is no longer possible.
In principle, end-to-end systems are open-vocabulary systems, when using appropriate modeling units, such as byte-pair encoded (BPE) \cite{sennrich2015neural} characters. However, in practice, words not seen during training are often not reliably recognized.
This is especially true for named entities. The reasons are that, a) the end-to-end network implicitly learns language model knowledge when being trained on transcribed speech data, and b) especially named entities often have a grapheme-to-phoneme relation that deviates from the general pronunciation rules of the language, as learned implicitly by the networks of the end-to-end system.

The recognition of words not seen during the training of an automatic speech recognition system, e.g., named entities, acronyms, or domain-specific special words, has been studied
%for a long time
in classical ASR systems\cite{suhm1993detection, suhm1994towards, maergner2012unsupervised, waibel2012system, waibel2015enhanced}.
Generally, the language model $P(Y)$ in equation \ref{eq:bayes} was modified.
%Finite state transducers
More recently, other works have combined statistical or neural language models with end-to-end ASR models using shallow fusion \cite{sriram2017cold, williams2018contextual, kannan2018analysis, huang2020class, kojima2022study}.
On the other hand, many recent works have used attention-based deep biasing
\cite{pundak2018deep, bruguier2019phoebe, jain2020contextual, huber2021instant, le2021contextualized, han2022improving, dingliwal2023personalization, huang2023contextualized, yang2024promptasr, sudo2024contextualized, yu2024lcb, xiao2025contextual,sudo2025owsm}. Some of them use only textual context information and some also include pronunciation information.
The problem with the former is that the model might not be able to relate audio and corresponding text of previously unseen words.
In that case the model is not able to recognize the previously unseen word and users have no effective way to correct this.
The problem with the latter is that such information is difficult to annotate by users.

In order to address this problem we 
1) propose a method (see Section \ref{sec:method}) that can take advantage of corrections of substitution errors provided during inference,
2) demonstrate that this method achieves a relative improvement between 22\% and 34\% in biased word error rate (BWER; see Section \ref{sec:metric}) compared to a text-based replacement method, while maintaining the overall word error rate, and
3) show that this method uses one correction of a substitution error more efficiently than the text-based replacement method.

%Supervised adaptation \cite{huber2020supervised}

%Implemented in \cite{huber2023end}

\section{Background} % Approach
\label{sec:background}

An auto-regressive end-to-end ASR model directly estimates the probability distribution
\begin{equation}
    P(Y_t|Y_0,\ldots,Y_{t-1};X)
    \label{eq:1}
\end{equation}
of the next token $Y_t$ given the already decoded sequence $Y_0,\ldots,Y_{t-1}$ and the audio input $X$.
The model is then used to find the word sequence $\hat{Y}$ with the highest probability
\begin{align*}
    \hat{Y}&=\argmax_{Y}P(Y|X)=\argmax_{Y}\prod_{t=1}^TP(Y_t|Y_0,\ldots,Y_{t-1};X),
\end{align*}
where $Y_0$ is the start of sequence token.

We work with a transformer-based encoder-decoder ASR model. First, $Y_0,\ldots,Y_{t-1}$ is embedded:
\begin{equation}
    E=Emb(Y_0,\ldots,Y_{t-1})\in\mathbb{R}^{t\times d}, \label{eq:emb}
\end{equation}
with $d\in\mathbb{N}$. Then, the decoder output is computed:
\begin{equation*}
O=Dec(H_X,E)\in\mathbb{R}^{t\times d},
\end{equation*}
where $H_X=Enc_{Audio}(X)$ is the encoded audio input. Finally, the output and softmax layers are applied:
\begin{equation*}
    \alpha=Linear(O)\in\mathbb{R}^{t\times n_{vocab}}, \label{eq:logits}
\end{equation*}
\begin{equation}
    p=Softmax(\alpha)\in\mathbb{R}^{t\times n_{vocab}}, \label{eq:lprobs}
\end{equation}
where $n_{vocab}$ is the vocabulary size.

Based on that model, we trained a context biasing model using the training scheme from \cite{huber2021instant} together with the architecture from \cite{sudo2025owsm}.
%In the following paragraphs, we describe the method:
%We describe the method as follows:
The method works as follows:
Equation \ref{eq:1} is replaced by
\begin{equation*}
    P(Y_t|Y_0,\ldots,Y_{t-1};X;Z),
\end{equation*}
where $Z$ is some context provided to the model. In our case,
\begin{equation*}
Z=(Z_1,\ldots,Z_L), L\in\mathbb{N},
\end{equation*}
is a list denoted context biasing list and each $Z_l$, $l\in\{1,\ldots,L\}$, is a word or short phrase the model is biased towards.

%\textbf{Context encoding:}
\subsection{Context encoding}

The model incorporates the context biasing list by first
% encoding each item with the
% %Whisper
% tokenizer and then the result is embedded using the embedding from the baseline model.
tokenizing and embedding each item. %with a tokenizer and then the result (denoted by $Z_l^{emb}$) is embedded.
% The result is denoted
% %by $Z_l^{emb}$.
% %\begin{equation*}
%     $Z_l^{emb} = Emb(Tokenize(Z_l))$.
% %\end{equation*}
Then,
an %the mBART-50 \cite{tang2020multilingual}
encoder is applied independently for each item, followed by a mean pooling %and afterward the mean
over the sequence dimension. %is computed.
This results in one vector per list entry:
%by $Z_l^s$,
\begin{equation*}
    Z^s = (Avg(Enc_{Context}(Emb(Tokenize(Z_l)))))_{l=1}^L.
\end{equation*}
%which can be interpreted as a summary vector of the list entry.

%\textbf{Context decoding:}
\subsection{Context decoding}

Then $Z^s$ is used to extend the vocabulary of the decoder. This is done by extending the output layer, which maps the output of the final decoder layer to the vocabulary, and extending the embedding layer.

In particular,
\begin{equation}
    \alpha_{Context}=\frac{Linear_2(O)\cdot Linear_3(Z^s)^T}{\sqrt d}\in\mathbb{R}^{t\times L}. \label{eq:acontext}
\end{equation}
is calculated and $\alpha$ in equation \ref{eq:lprobs} is replaced by
\begin{equation*}
    Concat(\alpha,\alpha_{Context})\in\mathbb{R}^{t\times (n_{vocab}+L)}.
\end{equation*}
Furthermore, $Y_0,\ldots,Y_{t-1}$ is replaced by $Y'_0,\ldots,Y'_{t-1}$, where $Y'_0,\ldots,Y'_{t-1}$ is calculated by replacing all subsequences of $Y_0,\ldots,Y_{t-1}$ which correspond to a context biasing list entry $Z_l$ with a dynamic token $v_l$.
Finally, $E$ in equation \ref{eq:emb} is replaced by
\begin{equation*}
E'=Emb(Y'_0,\ldots,Y'_{t-1}),
\end{equation*}
where dynamic tokens $v_l$ are embedded by $Linear_4(Z_l^s)$ and the rest of the tokens is embedded using $Emb$.

\subsection{Training}

During model training, in each step, the context bias list $Z$ is sampled from the labels of the corresponding batch. Specifically, we used a batch size of 16 and sampled on average three context biasing list entries per utterance of the batch. Then the context biasing list is filled up to a length of 200 with distractors sampled from other batches. 

% Therefore, for each token $Y_t$ the ground truth item $G_t\in\{0,\ldots,L\}$ in the context biasing list is known, where $G_t=0$ means that for that token no relevant information is available in the context biasing list. 
% The $loss$ consists of two cross-entropy parts.
% The first term classifies the next token $Y_t$
% of the sequence
% and the second term guides the network towards relevant context biasing list items:
% \begin{align*}
%     loss = &\ \frac{1}{T}\sum_{t=1}^TCE(f_\theta(Y_0,\ldots,Y_{t-1};X;Z),Y_t)\\
%     &+ \frac{\lambda}{MT} \sum_{m=1}^M \sum_{t=1}^TCE(S_t^m,G_t),
% \end{align*}
% \noindent where $f_\theta(Y_0,\ldots,Y_{t-1};X;Z)$ is the network output,
% $CE$ the cross-entropy loss function and $\lambda>0$ a hyperparameter.

\section{Method}
\label{sec:method}

The context biasing model (see Section \ref{sec:background}) learns during training to relate words in the context biasing list and the corresponding audio.
This works well and during inference the model can generalize to new words not seen during training\cite{huber2021instant, sudo2025owsm}.
%However, if there is a mismatch between pronunciation and orthography compared to what was learned during training, the model may not be able to recognize such words.
However, if this fails and the model is not able to relate audio and corresponding text, e.g. when there is a mismatch between pronunciation and orthography compared to what was learned during training, the model is not be able to recognize such words.

An example of the Yodas test set (see Section \ref{sec:data}) in which this is the case can be seen in Figure \ref{fig:approaches}.
Both the audio features $X_1$ and $X_2$ contain the named entity "Lottia" in the corresponding reference transcript.
However, the context biasing model is not able to recognize that (see Figure \ref{fig:approaches}, top).
In the first utterance the word "Lodea" is recognized, in the second utterance the word "Latia" is recognized.
Therefore, a text-based replacement of e.g. Lodea$\rightarrow$Lottia would not work for the first utterance. The same is shown quantitatively in Section \ref{seq:results}.

To deal with that problem, we noticed that for the words we are interested in (named entities, acronyms, and domain-specific special words) most of the time a substitution error occurs
%(e.g. for both LibriSpeech test sets $>$90\% of the time some word is aligned with the rare word as defined in Section \ref{sec:data}).
(the results in Section \ref{seq:results} show that more than 84\% of errors can be resolved by a substitution).
Let $Z_1$ be the word that should had been recognized and $\tilde{Z}_1$ the wrongly recognized one.
When we add $\tilde{Z}_1$ to the context biasing list and run the model again, we observe that the model mostly predicts the token of $\tilde{Z}_1$.
Therefore, the idea is to use the summary vector of $\tilde{Z}_1$ (instead of $Z_1$) in equation \ref{eq:acontext} but keep using $Z_1$ for $E'$.
We denoted this approach as context biasing + replacement and use a context biasing list entry $\tilde{Z}_1\rightarrow Z_1$ for illustration purposes (see Figure \ref{fig:approaches}).

In practice, our method would be applied as follows: Before running the model, a context biasing list can be supplied that contains words (named entities, acronyms, or domain-specific special words) that are likely to occur in the speech.
While running the model to recognize speech, users can correct substitution errors of important words and add them to the context biasing list.
%While running the model to recognize the speech, substitution errors of important words can be corrected by users and added to the context biasing list.
Our method then improves using these corrections (as shown in Section \ref{seq:results}).

\section{Experiments}

\subsection{Data}
\label{sec:data}

To evaluate our method, we created a test set from the English data of the Yodas data set\cite{li2023yodas}.
%Similarly to the method used in \cite{le2021contextualized}, we extracted the rare words of the Yodas data set references that occur at least four times but only in one of the YouTube videos.
Our goal was to identify cases where a standard context biasing model consistently fails on the same rare words — a necessary condition for meaningfully assessing whether our method can improve.
%Similarly to \cite{le2021contextualized}, we extracted as rare words those words from the references of the Yodas data set that occur only rarely. Specifically, we extracted words that occur at least four times and only in one of the YouTube videos.
%(hoping that at least two are misrecognized such that we can use them to test our approaches).
Following \cite{le2021contextualized}, we define rare words as words appearing in the references that occur infrequently overall yet are specific to a single YouTube video. Concretely, we retained words occurring at least four times but exclusively within one YouTube video. This yielded 6363 utterances (47.8 hours of audio) containing 8510 rare word occurrences across 1360 unique rare words.
%This resulted in 6363 utterances, 47.8 hours of audio, 8510 rare words and 1360 unique rare words.
%Then, we ran our context biasing model on the result, where for each utterance exactly the rare words of that utterance were used as context biasing list.
For each utterance, we ran our context biasing model using exactly the rare words of that utterance as the biasing list, reflecting an oracle biasing scenario.
%Afterwards, we filtered for utterances in which at least one rare word was misrecognized and for rare words which are misrecognized in at least two utterances. This resulted in 300 utterances, 2.24 hours of audio, 379 rare words and 94 unique rare words.
To focus our evaluation on the cases most relevant to our method, we filtered this set to retain only utterances in which at least one rare word was misrecognized, and only rare words that were misrecognized in at least two distinct utterances. The resulting test set comprises 300 utterances (2.24 hours of audio) with 379 rare word occurrences spanning 94 unique rare words on which the baseline context biasing model reliably struggles.

%We also tried to evaluate our method on Earnings-21\cite{del2021earnings},  LibriSpeech\cite{panayotov2015librispeech}, Fleurs\cite{conneau2023fleurs} and Voxpopuli\cite{wang2021voxpopuli}.
%However, after applying the context biasing model and filtering for misrecognized rare words, the resulting test set was too small to obtain significant differences between the approaches we compare. For example, in the Earnings-21 test set, only 5 rare words were misrecognized at least twice.

We applied the same procedure to Earnings-21\cite{del2021earnings},  LibriSpeech\cite{panayotov2015librispeech}, Fleurs\cite{conneau2023fleurs} and Voxpopuli\cite{wang2021voxpopuli}. The resulting test sets to evaluate misrecognized rare words were too small to yield a statistically meaningful evaluation. For example, in the Earnings-21 test set, only 5 rare words were misrecognized at least twice.

% We evaluate our context biasing method on two test sets: Earnings-21\cite{del2021earnings} and LibriSpeech\cite{panayotov2015librispeech}. For Earnings-21 we extracted rare words from the given annotations. 
% % For the experiments in this paper we created a
% % \href{https://github.com/chuber11/earnings21-new_words-dataset}{earnings-21 testset}
% % %\footnote{The new-words dataset is available \href{https://github.com/chuber11/earnings21-new_words-dataset}{here}.}
% % based on the earnings call dataset Earnings-21 \cite{del2021earnings}.
% %We extracted words from the categories \emph{named entity} (of persons), \emph{acronym} (abbreviations), and \emph{domain-specific special word} (products, events, laws, locations, and organizations) from the annotations.
% %We extracted rare words from the given annotations.
% For LibriSpeech we followed \cite{le2021contextualized} and used as rare words all words in the reference that fall outside 10\% of the most common words of our training data.
% Then, we filtered those rare words that occurred in at least two utterances.
% This resulted in 637, 726 and 821 utterances containing at least one rare word each with a total length of 1.23, 1.91 and 1.91 hours for earnings-21, LibriSpeech test-clean and LibriSpeech test-other, respectively. In total, there are 251, 339 and 353 unique rare words, respectively.
% We also tried to evaluate our approach on Fleurs and Voxpopuli, however, for these datasets the number of utterance after filtering was too low for evaluation.

\subsection{Models}

We use Whisper \cite{radford2023robust} (whisper-large-v2) as
%the baseline ASR model.
our speech foundation model.
The context biasing list is tokenized / embedded using the Whisper tokenizer / Whisper embedding, and
the context is encoded using a transformer encoder (the encoder of mBART-50 \cite{tang2020multilingual}).
%and the mBART-50 encoder \cite{tang2020multilingual} is used for the context encoding.

We trained the context biasing model on Common voice \cite{ardila2019common}.
We only trained the context encoder and the added linear layers. In contrast to \cite{sudo2025owsm}, we did not train the embedding and output layer. This has the advantage to prevent catastrophic forgetting \cite{french1999catastrophic} of the representations learned by the baseline model. Since we do not have access to the training data of the baseline model, this approach yielded substantially better overall performance. 

%During decoding of the testsets, the context biasing list contains the rare words belonging to the utterance which is currently decoded.
%Optionally, we add up to 250 distractors chosen randomly from the other rare words of the testset.

During decoding of the test set, the context biasing list contains the rare words belonging to the utterance which is currently decoded.
%Optionally, we add all 93 distractors chosen from the other rare words of the test set.
Optionally, we add all other rare words from the test set to the context biasing list as distractors.

\subsection{Approaches}
\label{sec:approaches}

To generate the changed context biasing list for the approach context biasing + replacement, we manually annotated the substitution errors $\tilde{Z}_1\rightarrow Z_1$ of the approach context biasing for all rare words. This resulted in 228 and 226 replacements for the hypotheses without and with distractors, respectively.
%To generate the changed context biasing list for the approach context biasing + replacement, we ran the context biasing model with zero distractors.
%Then we annotated the substitution errors $\tilde{Z}_1\rightarrow Z_1$ for all rare words. % using a large language model (LLM) followed by manual correction.
%This step could be automated, for example, using a LLM.
Finally, we add to the rare words of an utterance the replacements $\tilde{Z}_1\rightarrow Z_1$ of $other$ utterances which contain the same rare word (see Figure \ref{fig:approaches}, middle). %By the construction of our Yodas test set this always exists.
%Note, that this is only done for rare words the context biasing model is not able to recognize, as otherwise our approach would not be used in practice.
Examples for the replacements are: Lodea$\rightarrow$Lottia, Latia$\rightarrow$Lottia, Röding$\rightarrow$Rekin, Röging$\rightarrow$Rekin, Lindstra$\rightarrow$Lenstra, Lunster$\rightarrow$Lenstra, PPAL$\rightarrow$PIPOW, PayPal$\rightarrow$PIPOW.

To investigate the effect the number of replacements has, we restrict the number of added replacements per rare word between 1 and 4 and randomly sample this number of replacements. We have no rare word with more than 4 different replacements. 
When not using distractors, we obtain for a maximum of 1, 2, 3 and 4 replacements per rare word in total 244, 344, 380 and 382 replacements, respectively.
When using distractors, we obtain for a maximum of 1, 2, 3 and 4 replacements per rare word in total 242, 340, 371 and 376 replacements, respectively.
%For a maximum of 1, 2, 3 and 4 replacements per rare word we obtain 244, 344, 380 and 382 replacements used in total, respectively.
The number of replacements is higher than 228/226 because one replacement can be used for multiple occurrences of the same rare word. 

%When adding multiple replacements for one rare word, we noticed a performance drop in BWER. We eliminated the problem by changing the output probabilities of the decoder. In particular, if in a decoding step a dynamic token has the highest score, we set the scores of the other dynamic tokens to minus infinity.

For comparison, we compare with two approaches:

% \noindent 1) Context biasing + replacement oracle:
% This approach is similar to context biasing + replacement; however, the replacements $\tilde{Z}_1\rightarrow Z_1$ for the context biasing list are not taken from all other utterances containing the same rare word but only from the $same$ utterance (see Figure \ref{fig:approaches}, bottom).

\noindent 1) Context biasing + text replacement:
For this approach, we do not run the context biasing model with a context biasing list containing replacements, instead we take the hypotheses of the approach context biasing and apply the respective replacements that are used in the approach context biasing + replacement on the hypotheses.

\noindent 2) Context biasing + text replacement oracle:
This approach is similar to the previous approach;
%however, the replacements that are used in the approach context biasing + replacement oracle are applied to the hypotheses
however, the replacements $\tilde{Z}_1\rightarrow Z_1$ for the context biasing list are not taken from other utterances containing the same rare word but only from the $same$ utterance
(see Figure \ref{fig:approaches}, bottom).

\subsection{Metrics}
\label{sec:metric}

The performance of an ASR system is typically measured using the word error rate (WER). To measure how well a context biasing method is working, \cite{le2021contextualized} extended this metric to UWER (unbiased WER measured on words not in the biasing list) and BWER (biased WER measured on words in the biasing list), given a test set together with a corresponding context biasing list.
%Due to space limitations we report BWER and WER only.
%We evaluate these metrics along with WER and F1-score to compare different approaches.
We evaluate these metrics along with WER to compare different approaches.

% Recently, PIER \cite{ugan2025pier}
% %, a recall based metric, 
% was introduced. We also tried this metric and found that
% for our use case,
% it is similar to BWER and UWER. %because for our approaches precision is quite high.

\begin{table*}[t]

\caption{Results on the Yodas test set: BWER/UWER/WER in \% for the different approaches depending on the maximum number of added replacements per rare word with and without adding distractors.
%The best results in the different blocks are underlined, printed bold or wavy underlined, respectively.
}

\begin{center}

%\vskip -10pt

\begin{tabular}{|c|c|c|c|c|c|}
\hline
Approach & \diagbox[width=2.5cm]{Distractors}{\makecell{Number\\repl.}} & 1 & 2 & 3 & 4\\
\hline
\makecell{Context biasing} & \ding{55} & \multicolumn{4}{c|}{82.8/6.4/7.8}\\
\makecell{Context biasing + text replacement} & \ding{55} & 46.2/6.0/6.8 & 36.9/6.0/6.6 & 34.6/5.9/6.5 & 34.6/5.9/6.5\\
\makecell{Context biasing + replacement} & \ding{55} & 30.6/6.0/6.5 & 27.2/6.0/6.4 & 26.9/6.0/6.4 & 26.9/6.0/6.4\\
\makecell{Context biasing + text replacement + replacement} & \ding{55} & 24.5/5.9/6.3 & 21.6/6.0/6.3 & 21.6/6.0/6.3 & 21.6/6.0/6.3\\
\makecell{Context biasing + text replacement oracle} & \ding{55} & \multicolumn{4}{c|}{13.2/5.7/5.9}\\
\hline
\makecell{Context biasing} & \ding{51} & \multicolumn{4}{c|}{83.6/6.6/8.1}\\
\makecell{Context biasing + text replacement} & \ding{51} & 47.0/6.3/7.1 & 38.3/6.2/6.8 & 35.9/6.2/6.8 & 35.6/6.2/6.8\\
\makecell{Context biasing + replacement} & \ding{51} & 34.3/6.2/6.8 & 28.0/6.2/6.6 & 27.7/6.1/6.6 & 27.7/6.1/6.6\\
\makecell{Context biasing + text replacement + replacement} & \ding{51} & 26.4/6.2/6.6 & 21.4/6.1/6.4 & 21.4/6.1/6.4 & 21.4/6.1/6.4\\
\makecell{Context biasing + text replacement oracle} & \ding{51} & \multicolumn{4}{c|}{14.5/6.0/6.2}\\
\hline
\end{tabular}

\label{tab:results}

\end{center}

\end{table*}

\section{Results}
\label{seq:results}

The results for the Yodas test set can be seen in Table \ref{tab:results}.

%Let's first look at the results without distractors.

By construction of the Yodas test set (see Section \ref{sec:data}) the BWER of the Context biasing approach (without distractors) is very high (82.8\%). Note, that the BWER is not 100\% because some utterances contain a rare word multiple times and not every instance of the rare word is misrecognized. On the other hand, the BWER of the approach Context biasing + text replacement oracle is only 13.2\%. Therefore, over 84\% of errors can be resolved by a substitution.

When comparing the approach Context biasing + text replacement (without distractors), we see that it is between 44\% and 58\% (relative) better than Context biasing. Furthermore, a higher number of replacements reduces BWER by 25\%.

Next, we see that the BWER of Context biasing + replacement is between 22\% and 34\% better than Context biasing + text replacement.
A higher number of replacements also helps this approach in terms of BWER, but only 12\% relative.
Together with the better performance for number replacements 1, this suggests that one given replacement per rare words is used more efficiently than in Context biasing + text replacement. We checked statistical significance using Bootstrap Resampling\cite{berg2012empirical} with one million samples and found that the difference (46.2\% vs. 30.6\%) is statistically significant with p-value $2.0e^{-6}$. The same holds for number replacements 4, where we have BWER 34.6\% vs. 26.9\% and a p-value of $3.9e^{-3}$. 

The computational overhead when adding replacements to the context biasing list is negligible. The context encoder output can be reused when decoding multiple utterances and extending the embedding / output layer is insignificant compared to the vocabulary size (which is around 250k).

The approach Context biasing + replacement outperforms Context biasing + text replacement in the cases where the model predicts for different utterances containing the same rare word different words instead of the rare word. In this case, the text replacement does not match.
Examples include: Lottia (genus of sea snails), Rekin (name of a festival), Qama (name), Finotex (company), BANI (concept), Chariklo (planet), Lenstra (name), Parasuram (person name), PIPOW (framework name), Kirima (name).

Combining the methods replacement and text replacement delivers even better results: BWER is between 38\% and 47\% better than text replacement alone. Furthermore, up to 88\% of the errors (which can be corrected by the oracle approach) can be corrected by the combination of replacement and text replacement. The oracle approach with BWER 13.2\% is still better than that, however, it has access to oracle information.

Looking at the performance with added distractors, we see the same behavior compared to not adding distractors but mostly with slightly lower performance in all metrics, which is expected. Comparing the approaches Context biasing + replacement and Context biasing + text replacement, we have for number replacement 1 (47.0\% vs. 34.3\% BWER) and 4 (35.6\% vs. 27.7\% BWER) p-values of $1.6e^{-4}$ and $7.3e^{-3}$, respectively.

Finally, the UWER performance changes less than 2\% relative (excluding the Context biasing and oracle approaches) while the WER performance improves up to 7\% between text replacement, replacement and the combination of both because the BWER performance improves.

\subsection{Limitations}

Our proposed approach context biasing + replacement can only be applied if there is a substitution error, not if there is a deletion error.
Furthermore, if the substitution error produced a word with a very high number of occurrences, our approach likely will produce false positives.
%Therefore, it might be better not to use the replacement approach or to keep the word in the context biasing list only for the relevant session and then transfer the knowledge through continuous learning \cite{huber2025continuously}.
For such cases, it might be best to keep the replacement in the context biasing list only for the relevant session and then transfer the knowledge through continuous learning \cite{huber2025continuously}.

We also tried to generate the replacements automatically from the utterances where the context biasing model was able to correctly recognize a rare word compared to the baseline model which failed to do so. However, this did not lead to improvements, suggesting that manual corrections are necessary.

%TODO: Tried to learn using replacements which the model produced (without human annotation): does not work well?

% Instead of boosting the tokens of the context biasing list item we also tried to replace the probabilities of the next token prediction with the correct one-hot vector.
% However, that did not work.
% The WER for 0 and 250 distractors jumped to 15.7\% and 17.7\%, respectively. TODO: remove this paragraph?
% TODO: explain why?

%{'name': 'add inf, distr 0  , beam1   ', 'bwer': 12.54, 'uwer': 13.36, 'wer': 13.25, 'add': 'inf', 'distractors': 0}
%{'name': 'add inf, distr 0  , beam4   ', 'bwer': 14.66, 'uwer': 15.85, 'wer': 15.69, 'add': 'inf', 'distractors': 0}
%{'name': 'add inf, distr 250, beam1   ', 'bwer': 14.6, 'uwer': 14.89, 'wer': 14.85, 'add': 'inf', 'distractors': 250}
%{'name': 'add inf, distr 250, beam4   ', 'bwer': 16.95, 'uwer': 17.79, 'wer': 17.68, 'add': 'inf', 'distractors': 250}

\section{Conclusion}

%In this work, we addressed the challenge of recognizing words with a pronunciation-orthography mismatch in ASR systems. While existing context biasing methods improve recognition of unseen words, they might struggle when the pronunciation of a word deviates from its orthography and have no effective way to learn from corrections.

In this work, we addressed the challenge of recognizing words where existing context biasing methods are not able to relate audio and corresponding text, e.g. in case of a pronunciation-orthography mismatch.

We proposed a method, context biasing + replacement, that allows corrections of substitution errors provided during inference to enhance recognition accuracy. By incorporating these corrections into the context biasing list, our method significantly improves the recognition of such problematic words. Our experiments demonstrated a relative improvement in BWER between 22\% and 34\% compared to a text-based replacement, while maintaining the overall word error rate. Our method can use one correction more efficiently compared to the text-based replacement.

\section{Acknowledgment}

\ifcameraready
This research was supported in part by a grant from Zoom Video Communications, Inc.
Furthermore, the projects on which this research is based were funded by the Federal Ministry of Education and Research (BMBF) of Germany under the number 01EF1803B (RELATER),
the Horizon research and innovation program of the European Union under grant agreement No 101135798 (Meetween),
and the KIT Campus Transfer GmbH (KCT) staff in accordance with the collaboration with Carnegie – AI.
The authors gratefully acknowledge the support.
\else
BLIND
\fi
%This research was supported by Zoom Video Communications, Inc., the Federal Ministry of Education and Research (BMBF) of Germany under the number 01EF1803B (RELATER), the Horizon research and innovation program of the European Union under grant agreement No 101135798 (Meetween) and the KIT Campus Transfer GmbH (KCT) staff in accordance with the collaboration with Carnegie – AI.

\section{Use of generative AI tools}
Generative AI tools were used in a limited capacity during the preparation of this work. Specifically, AI-assisted code completion was employed to support software development tasks. Language model suggestions were used to refine the clarity and style of the written text. Additionally, generative AI tools assisted in the enhancement of figures. All substantive intellectual contributions, including the research design, methodology, analysis, and conclusions, are entirely the authors' own.

\bibliographystyle{IEEEtran}
\bibliography{mybib}

\end{document}